# VeriMedi: Pill Identification using Proxy-based Deep Metric Learning and Exact Solution

Tekin Evrim Ozmermer*, Viktors Roze
Stanislavs Hilcuks, Alina Nescerecka

Accenture Latvia ATC, Latvia
*tekin.evrim.ozmermer@accenture.com

**Abstract.** We present the system that we have developed for the identification and verification of pills using images that are taken by the VeriMedi device. The VeriMedi device is an Internet of Things device that takes pictures of a filled pill vial from the bottom of the vial and uses the solution that is presented in this research to identify the pills in the vials. The solution has two serially connected deep learning solutions which do segmentation and identification. The segmentation solution creates the masks for each pill in the vial image by using the Mask R-CNN model, then segments and crops the pills and blurs the background. After that, the segmented pill images are sent to the identification solution where a Deep Metric Learning model that is trained with Proxy Anchor Loss (PAL) function generates embedding vectors for each pill image. The generated embedding vectors are fed into a one-layer fully connected network that is trained with the exact solution to predict each single pill image. Then, the aggregation/verification function aggregates the multiple predictions coming from multiple single pill images and verifies the correctness of the final prediction with respect to predefined rules. Besides, we enhanced the PAL with a better proxy initialization that increased the performance of the models and let the model learn the new classes of images continually without retraining the model with the whole dataset. When the model that is trained with initial classes is retrained only with new classes, the model's accuracy increases for both old and new classes. The identification solution that we have presented in this research can also be reused for other problem domains which require continual learning and/or Fine-Grained Visual Categorization.
**Keywords:** Image Classification, Deep Metric Learning, Fine-grained Visual Categorization, Pill Identification, Continual Learning

# 1. Introduction

Medication therapy is an important part of the health system. According to Agency for Healthcare Research and Quality [5], the process before the patient receives drugs, consists of several steps, which include the selection of the appropriate medication and plan by the clinician, interpretation of the prescription by the pharmacist, checking drug interactions and patient's allergies, and, finally, the release of the correct drug in the appropriate quantity to a patient. The errors in any of these processes can have negative consequences for a patient's health. Therefore, healthcare professionals should comply with the "Five Rights" of medication safety which are "administration of the right medication, in the right dose, at the right time by the right route, to the right patient" [5]. Because of the large number of different pills and the similarity of medication names and their appearance, medication errors can occur in the dispensing process and become a risk factor for adverse drug events [5].

While it is impossible to eliminate errors arising from human decisions making, which arise from human-driven decisions, some routine and purely technical actions can be improved by using digital and automation solutions. For example, using electronic prescriptions can prevent errors related to the interpretation of the prescription, and automated dispensed devices can control the release and distribution of the drugs. Computer vision (CV) is another promising technique that can improve and secure the drug administration process in pharmacies.

The identification of pills by the appearances of the pills is a challenging subject because the appearances of the different pill types are similar to each other. Pills can be identified by a person by jointly using the imprint, shape, and color information of the pill. While some pills have similar shapes and colors and differ only by their imprints, some pills have the same imprint and differ from each other by shape and color information. There are some cases where the imprints of the pills differ from each other just by one characteristic. Therefore, the pill identification problem is a part of Fine-Grained Visual Categorization.

The need to develop pill recognition solutions led to several challenges. In January 2016, the pill recognition challenge was announced for the development of a system to identify the pills by using the images of the pills. The dataset contained 2000 reference images and 5000 consumer quality images of 1000 pills. The second dataset that was used for segregation testing contained 1000 pills [6]. The winners of the challenge developed systems that can identify consumer quality images with average precision scores of 0.27, 0.09, and 0.08. For each winner solution, the correct images were in 43%, 25%, 11% of the images that are retrieved by the model from query/consumer images [6]. This is a considerable achievement, considering the fact that the retrievals are made from 5000 query/consumer images [6]. However, the solutions that are made for the challenge cannot continuously learn. This is an important fact because the number of pill types changes every year with the release of new drugs.

In September 2020, Microsoft has announced another challenge and provided a dataset with 13k images representing 9804 appearance classes (two sides for 4902 pill types) [7]. The dataset came with the challenge of few-shot learning because of the number of appearances for each pill type. Moreover, the dataset contained a high number of pill types and their appearance was similar [7]. A Convolutional Neural Network (CNN) model using softmax cross-entropy loss (plain classification model) and Deep Metric Learning model with the combination of four different loss functions (multi-head metric learning) were investigated in the study. As a result, plain classification models performed poorly, while the multi-head metric learning models have achieved over 95% mean average precision and 90% global average precision at first retrieval [7].

To reduce the pill identification errors, speed up the identification process and prevent physical contact between pills and pharmacist, two matters are taken into account:

1. The identification and verification process should be done without taking the pills out of the vial that are in.
2. To avoid the illumination and pose challenges that are mentioned in [6, 7], it is necessary to create different poses and illumination conditions so that more information can be gathered from the pill appearances.

To satisfy the matters mentioned above, the VeriMedi device [1] was developed.

The goal of this research was to design and develop a pill identification system that can predict pill types from noisy images with multiple same type pills that were introduced to the system before, verify or deny the given predictions to achieve zero false-positives, avoid giving verified predictions for unknown objects and pill types. The system also should be able to predict the new set of pill types that are introduced to the system in addition to the previous set of pill types, which means that the system should be able to learn continuously.

First, the case study description is given. Second, datasets that were prepared in this research are discussed. Then, pill segmentation and identification solutions are explained. After that, the experiments that were conducted to make decisions for the model architecture, loss function, and classifier algorithm are given along with the demonstration of the methods that were proposed in this research. Finally, the results of the final solution are given.

## 2. Case Study Description

This research is a part of the VeriMedi project, which is aimed to develop a system for the prescription verification process in pharmacies. The project can be broadly divided into hardware and software development. The hardware and the control of hardware are designed and developed by MindTribe (part of Accenture Industry X in North America). VeriMedi is a physical device that is designed specifically for the project and patented by Accenture in 2020 [1]. It appears as VeriMedi device later in the text.

In turn, this study is about a software that detects, then recognizes pills from the images acquired by the VeriMedi device, and records the results. Both hardware and software were designed so that the system could predict the pills and verify the prescription without opening the lid and without removing the pills from the vial.

VeriMedi device's main parts are Raspberry Pi, Raspberry Pi camera, light sources, and a robotic arm (Figure 1). The camera is located under the transparent glass platform and surrounded by light sources forming a circular shape. The camera objective is turned up to capture the image of the bottom of the pill vial which is placed on the glass platform.

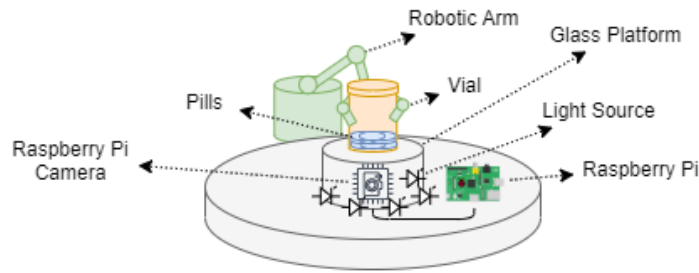

*Figure 1: A basic illustration of the VeriMedi device*

The overall process is displayed in Figure 2. The filled vial is put on the platform of the VeriMedi device and the process is started by the pharmacist. The Raspberry Pi takes 7 images with 7 different light conditions of the pill vial bottom with one or multiple pills and measures weight. Images and a request ID are sent to the analysis software that is located in the cloud. The services make the analysis and send the prediction, request ID, and verification status back to the Raspberry Pi. Raspberry Pi counts the number of pills in the bottle by using total weight and pill type information. The results are displayed on the VeriMedi device. If verification fails, the device shakes the bottle with an artificial arm controlled by servo motors to distribute pills differently within the vial. When a new request with the same pills having different distribution is made to the services, request ID is used so that the system can aggregate not only the predictions of current pills but also the predictions of previous pills.

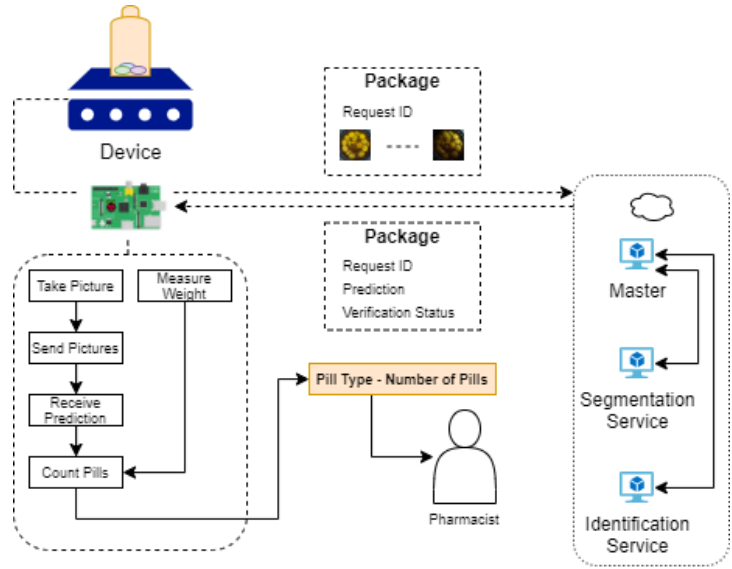

*Figure 2: Overall Solution Diagram.*

# 3. Datasets

To develop and assess the solution, two different datasets were created. The first dataset is a ShakeNet dataset, acquired with the VeriMedi device. Due to the limited number of pill types in the ShakeNet dataset, an additional dataset that has a wider range of pill types was generated. SyntheticNet was generated by locating reference pill images from NIH [6] dataset inside the empty vial images that were captured by the VeriMedi device.

## 3.1 ShakeNet Dataset

Image data is acquired by the VeriMedi device in a total of seven images for a batch, where one of the images was taken with all light sources on, and six images were taken with one of the light sources on (Figure 3). Collecting images with different illumination allowed enhancing the contrast of pill features such as imprints and shape in the image. The size of acquired images is 1944 x 2592, and the image format is JPEG.

ShakeNet dataset consists of 1400 raw vial images, representing 20 pill types. There were 10 poses for each class where each pose represents a different distribution of the pills in the pill vial. Each pose had 7 light conditions.

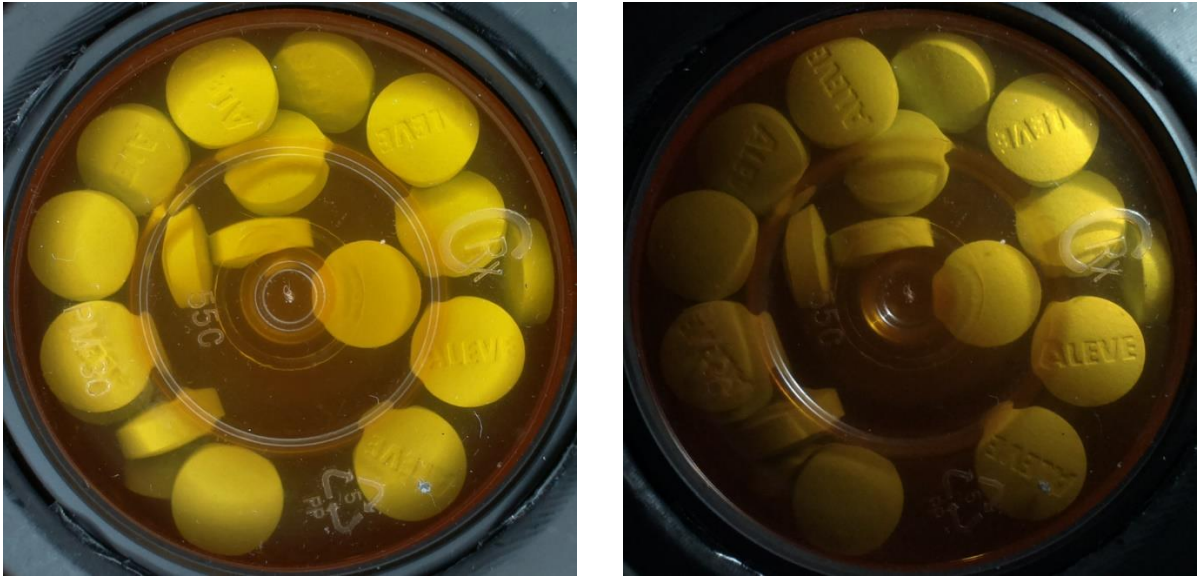

*Figure 3: Images with different light sources. On the left, it is a fully illuminated image with all light sources on. On the right, it is a partially illuminated image with only one light source on it.*

The raw image represents the pill vial bottom, where the appearance of the pills is affected by vial color, engravings, recessed base, overlapping of the other pills (Figure 4).

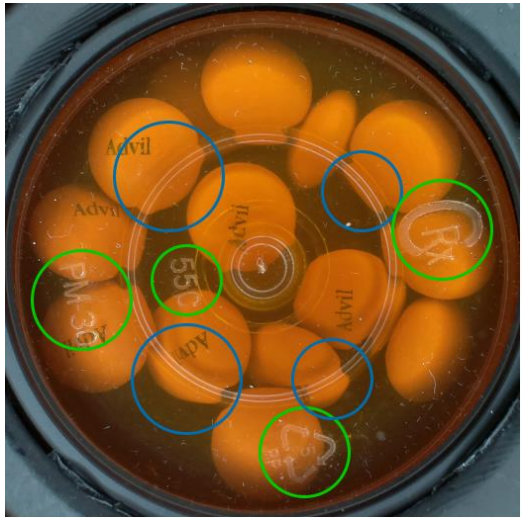

*Figure 4: Raw image from the VeriMedi device. Blue circles indicate the deformations caused by the recessed base. Green circles indicate the engravings on the vial.*

### 3.2 SyntheticNet Dataset

SyntheticNet is a dataset that was synthetically generated to imitate the images acquired from the VeriMedi device. The generation process has three inputs which are single pill image, vial image, and valid region mask.

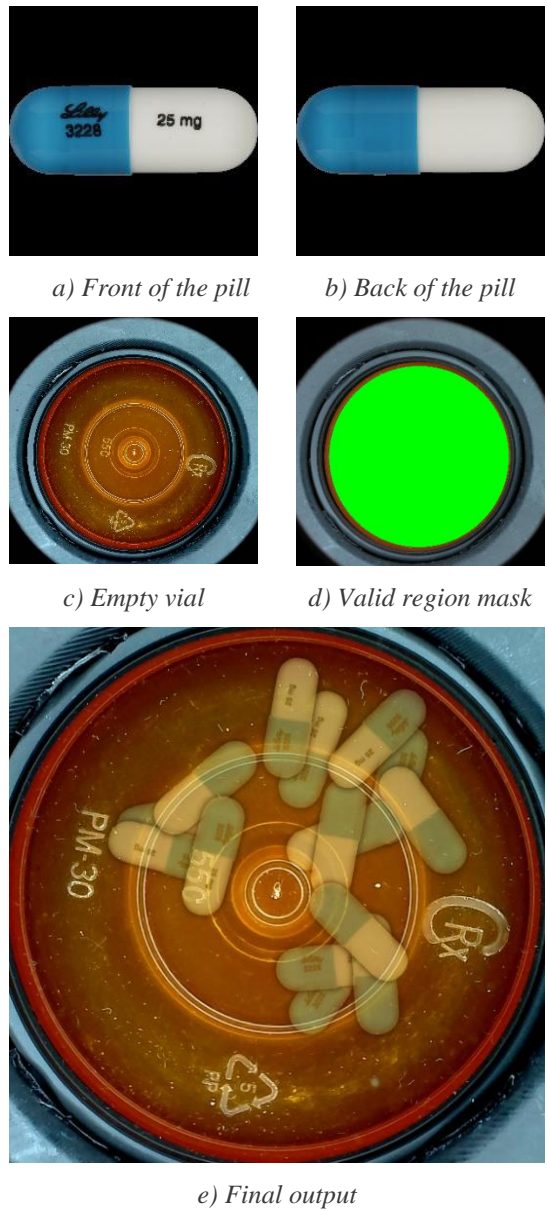

*a) Front of the pill*    *b) Back of the pill*

*c) Empty vial*    *d) Valid region mask*

*e) Final output*

*Figure 5: Inputs for image generation for SyntheticNet dataset and the output image.*

The dataset contains 100 pill types with 10 poses for each class. Each image takes approximately 13 seconds to be created by using the CPU of Dell Precision 5540. The programming language is Python. The Python libraries that are used in the code are Sci-Kit Image [12], TorchVision [13], PIL [14], SciPy [15], NumPy [16]. Single pill images are imported from an open-source dataset [6].

The steps of the generation process are given below:

1. Import the single pill images of a pill type from NIH [6] pills. The pills are masked and have a black background. There are two images per single pill type, which are the front and back of the pills. The front images are labeled with 0 and the back images are labeled with 1 in the file name. When a pill is chosen to be the candidate to be located in a vial image, a random choice is made between the front and back of the pill.

2. Make a random choice to choose one of the empty vial images from several samples that were previously captured. Import the chosen vial and the valid region mask.
3. Generate a mask for the single pill image and bind the mask and pill image to each other so that the transforms that will be applied to one of them will affect both.
4. Calculate and create the central location circle (Figure 14) to define the candidate locations to place the pills. This process is made to embed the pills randomly in a predefined boundary so that the pills will not look like they are out of the vial.
5. Apply random rotation to the single pill image and its mask.
6. Apply Gaussian Blur to the mask of a single pill image to create a shadow-like effect.
7. Embed the transformed single pill image in a predefined random location inside the boundary of the central location circle.
8. Update the central location circle by extracting the mask of the pill from the central location circle to avoid pills that are located on each other. If the area of the new central location circle is smaller than 10% of the area of the initial central location circle, cancel the addition of the next pill.
9. Repeat the process a maximum of 10 times. If the process is canceled by step 8, do not repeat the process.

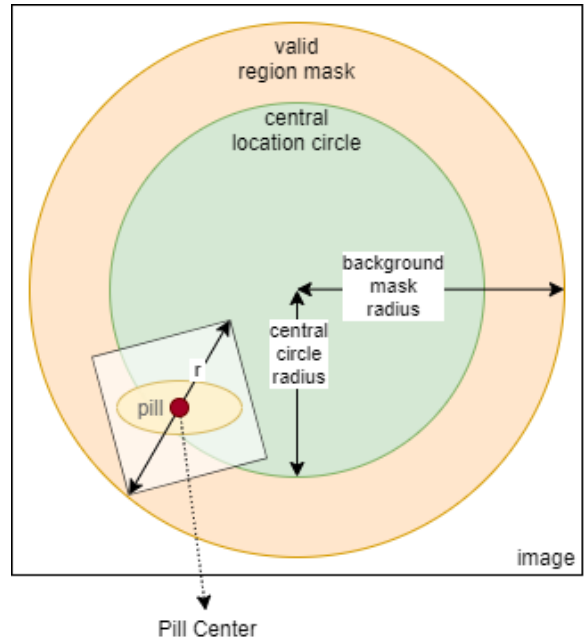

*Figure 6: Pill embedding area representation. Valid region mask is the manually labeled area, which corresponds with the pill vial bottom. The central location circle defines the area for the pill centers location.*

The radius of the central location circle was calculated by formula 2.

$$r_{clc} = r_{vrm} - (s_r^2 + s_c^2)^{\frac{1}{2}} \qquad (2)$$

*where:*
$r_{clc}$: *radius of central location circle*
$r_{vrm}$: *radius of valid region mask*
$s_r, s_c$: *row and column sizes of single pill image*

When the process of verification starts, a total of seven images are taken. One of the images is taken with all light sources are on, and six images are taken with one of the light sources is on.

## 4. Pill Segmentation

To avoid mispredictions that could be caused by noises from the pill bottom (See Figure 4), we evaluated the content type of the input images for pill identification.

| | | |
|---|---|---|
| Multiple Single Pills | Advantages | It contains only relevant features such as imprint, color, shape |
| | Disadvantages | It may not contain all features because of noise created by the imprints on the vial and the ribbed base of the bottle. It may not contain imprint information because the pill has two sides and the imprint of the pill is only on one-side. |
| One Whole Vial | Advantages | It contains all information that a pill type can have. |
| | Disadvantages | Contains all occurring noises which may cause misprediction or increase training complexity. |

*Table 1: Advantages and disadvantages of using whole vial images and multiple single-pill images.*

Using the information in Table 1, we decided to continue with multiple single pill images instead of single whole vial images because we want to avoid noisy data as much as possible. Therefore, the localization and segmentation of individual pills from the image became necessary. To achieve this goal, we applied pre-processing to the raw images that are captured by the VeriMedi device, labeled a portion of images from both ShakeNet and SyntheticNet datasets, and trained a segmentation model (Figure 7).

For the preprocessing, the vial appearance was cropped (in square shape) from the raw image to fit the edges of the vial to the edges of the image. After that, the image was resized to 1024x1024 so that it could be fed into the segmentation model for both training and inference.

For training the segmentation model, 82 whole vial images from both ShakeNet and SyntheticNet datasets were labeled manually using the VIA labeling tool [9]. Labels contained polygons labeled with the pill types, imprint polygons labeled with imprint types such as white pill imprint, black pill imprint, and polygons labeled with vial engravings.

For the segmentation model, we chose the Mask R-CNN model [2] and forked the open-source implementation of it from [17]. After training the segmentation model with the labeled data, we applied a background effect to the segmented pills. The blurred background type was chosen after experimenting (See Section 6.2) with different background types to understand which background type is most suitable.

When the training process of the segmentation model was done, it was used to create the dataset for the pill identification model. The multiple images were saved in a folder having the same name as the whole vial images that were used as the input (Figure 7).

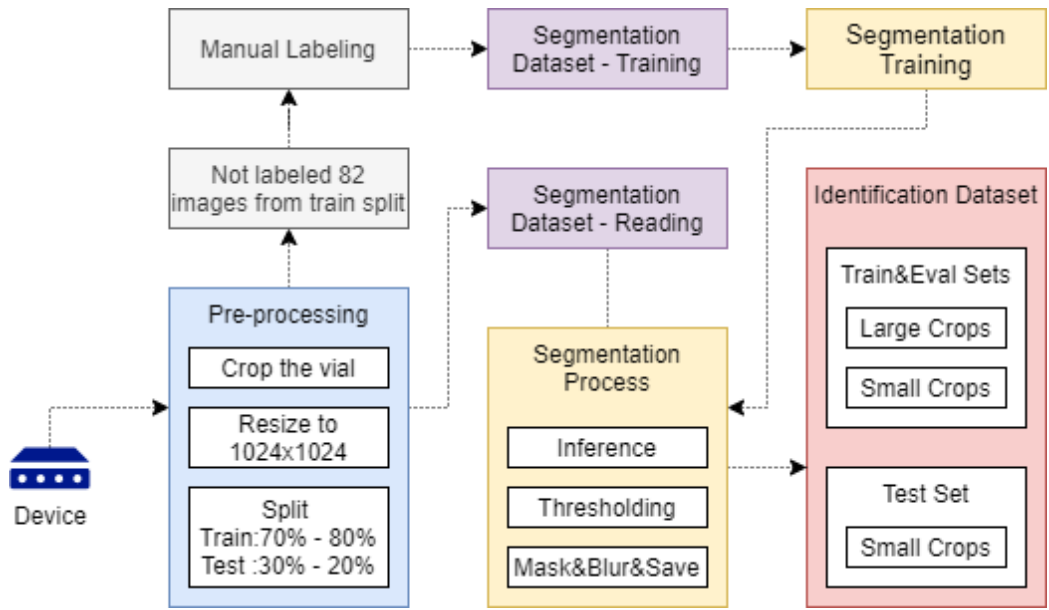

*Figure 7: Pill segmentation pipeline in development*

# 5. Pill Identification

After the segmentation process is executed, the class of every single pill that is cropped and processed should be predicted. This process is done with a pill identification solution.

Even though classic multi-class classification models have proven themselves in classification tasks, they must be retrained from scratch when a new set of classes are requested to be predicted. Because the number of pill types changes over time, a classic multi-class classification model cannot be used in our task.

On the other hand, a Deep Metric Learning (DML) model that is trained with a pair-based loss function extracts high-level features that can help us distinguish different classes from each other. The embedding vectors which are generated by the DML model can be saved and used later as a reference to predict the class of embedding vector that is queried during inference. Because the model has learned to extract features rather than mapping input to a specific output, new classes can be introduced to the model just by generating their embedding vectors and appending them to embedding vector collection to be used in inference later. Therefore, a DML model that is trained with a pair-based loss function can be used in our task.

At the same time, a DML model that is trained with proxy-based loss function learns to map input images to sparsely distributed embedding vectors. Even though this DML model may not be able to extract meaningful features as the DML model that is trained with pair-based loss function, which can help us later to predict new classes, it can still work with new classes if the number of proxies in the proxy set is increased.

Because DML models can enlarge the number of pill types that they can predict without retraining the model from scratch, DML was chosen to be used for the identification solution.

To use the DML, a backbone model that generates embedding vectors and a classifier algorithm that would be used to classify the embedding vectors were required (Figure 6).

During development, the single-pill images in the training set were fed into the trained DML model to generate embedding vectors. These embedding vectors were used to train a classifier. Then, a test was conducted to evaluate the performance of the overall identification solution.

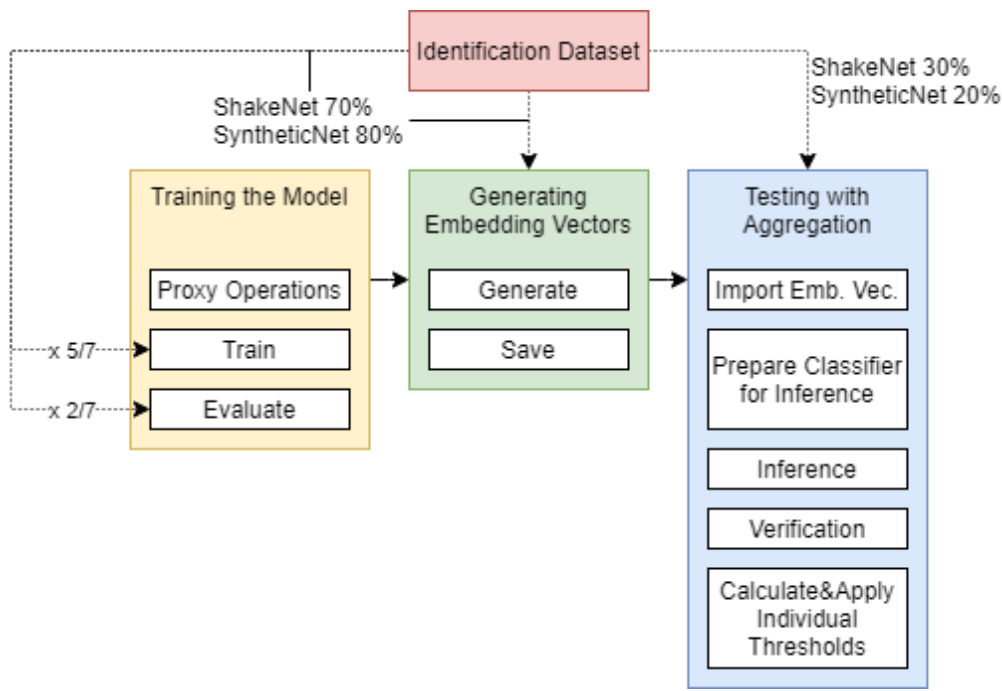

*Figure 8: Pill Identification pipeline using a DML model in the development*

## 5.1 Backbone Architecture

ResNet-34 architecture [11] was used as the backbone of the DML model. The architecture was chosen according to the conducted experiment (See Section 6.3). The model architecture can be seen in Figure 9. The model was forked from the GitHub repository of [3].

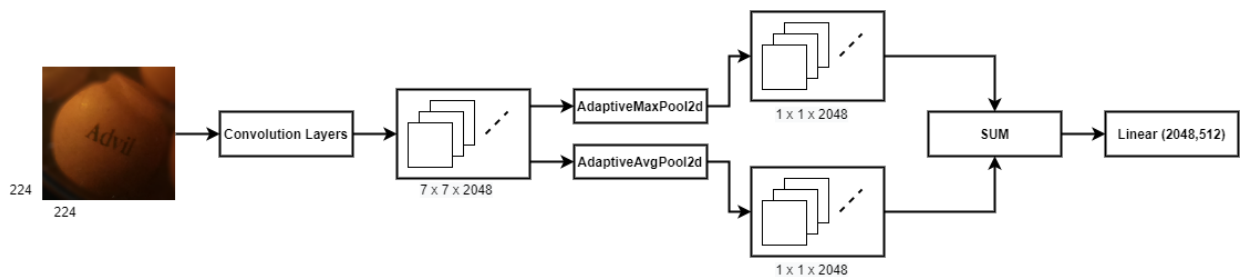

*Figure 9: The model architecture for the DML model. The latent vector that is generated by the convolution layers has the size 7x7x2048. After the convolution layers, adaptive max pooling and adaptive average pooling are applied to the latent vector separately. The resulted vectors are summed and fed into a fully connected layer outputs the embedding vector having size 512.*

The size of embedding vectors that are generated by the backbone model is crucial in DML. According to [3], a higher embedding size can give higher accuracy when proxy-based loss functions are used. On the other hand, the embedding size depends on the fully connected layers of the DML model. In the ResNet architectures, the second last layer has a size of 2048 and each one node that is added to the output layer (the layer representing the embedding vector) increases the number of parameters in the model by 2048.

Because the response time of the overall solution is an important factor, the optimal size of the embedding vector is determined as 512.

## 5.2 Loss Function for Deep Metric Learning

The loss function that is used to train a DML model is a crucial factor for creating a successful solution. While the pair-based loss functions can capture fine-grained features, proxy-based loss functions can converge to low error rates faster [3].

The problem domain of pill identification is defined as "Fine-Grained Visual Categorization (FGVC)" in [7]. Therefore, it was likely to use one of the pair-based loss functions. During the research, we used several different pair-based loss functions which are Contrastive Loss [20], Triplet Loss [21], Multi-Similarity Loss [22]. One of the challenges that we faced was the mining of the data. The pairs that are selected in the image batch are important to train the model to capture "correct" features while keeping the training complexity low. It was observed that the models can converge to low loss values during the training but make a considerable amount of mispredictions during the tests especially for the pill types having similar shapes and colors. This showed that the pair-based loss functions were leading the model to focus on shape and color more than the imprint while the imprint is the most distinguishing feature to make predictions. After this issue, we investigated our dataset more in-depth. It was seen that the pill types have different appearances on each side which have only shape and color features in common. This situation was leading the model to focus on shapes and colors while assuming the imprints as coincidental patterns. This required us to find a solution to solve the problem. We came up with two ideas which are given below:

1. Label the sides of the pills with different labels such as "Advil-Front" for the front-side of the Advil and "Advil-Back" for the backside of the Advil instead of labeling both sides as "Advil".
2. Use a proxy-based loss function instead of a pair-based loss function.

When we tried the first idea, the labeling process took a considerable amount of time. Labeling the same type of pills with different labels would also slow down the new pill registration process for the client at the production-level.

In theory, a model that is trained with pair-based loss function can extract meaningful features so that the features are stored to query them later. On the other hand, a model that is trained with a proxy-based loss function generates a representative vector of the input image rather than representative features. The reason for it is because pair-based models will learn to compare input samples while proxy-based models will learn to map the input to associated proxy vectors. On the other hand, proxy vectors that the model generates for the new class of images that the model is not trained with before may not be decomposed enough to distinguish them from the vectors of existing classes. Therefore, a new method that will decompose the proxies of the new classes is required if it is decided to choose a proxy-based loss function.

There are two types of proxy-based loss functions which are Proxy Anchor Loss (PAL) [3] and Proxy-NCA Loss [23]. PAL outperformed all other pair-based and proxy-based loss functions [3]. The performance of the PAL function in the benchmark datasets convinced us to choose it to use in our solution. To handle the decomposition problem of the proxies, Proxy Operations were designed and developed in this research.

*5.2.1 Proxy Operations*

In Proxy-Anchor Loss [3], proxies were defined randomly. According to the code given in the repository of [3], the proxies are defined with Kaiming Normal Distribution [11]. This creates a situation where proxies are not decomposed enough from each other, especially if the number of classes is close to or more than the embedding size (Figure 11).

Because our problem domain is "Fine-Grained Visual Categorization" [7] where some of the pills differ from each other only by one line or character, the DML model can generate similar embedding vectors for different pill types having similar appearances. To avoid this issue, proxies that represent each class should be defined as sparse as possible to each other so that the classifier will be able to distinguish the embedding vectors also for the pill types that have similar appearances to each other. Therefore, we proposed a method to decompose the proxies while they are being created. This method was used in three operations which are proxy creation, proxy addition, proxy enhancement.

The model training process using the proxy operations is shown in Figure 10. Firstly, the proxy creation operation is used to define initial proxies. When a new set of classes arrive, the decomposition for new proxies starts. The images of the new classes are fed into the pre-trained model to generate embedding vectors. The mean of embedding vectors of each class is calculated to create one candidate proxy for each class. These candidate proxies are concatenated with initial proxies and sent to the proxy addition operation. During proxy addition operation, only the new proxies are registered as parameters to be updated. Decomposition is applied to the new proxies to maximize the distance between each other and existing proxies. Then, the proxy enhancement operation starts. In proxy enhancement, all proxies are registered as parameters to be updated. Then, the decomposition is applied to all proxies. After that, the enhanced proxies are registered as proxies for the PAL function.

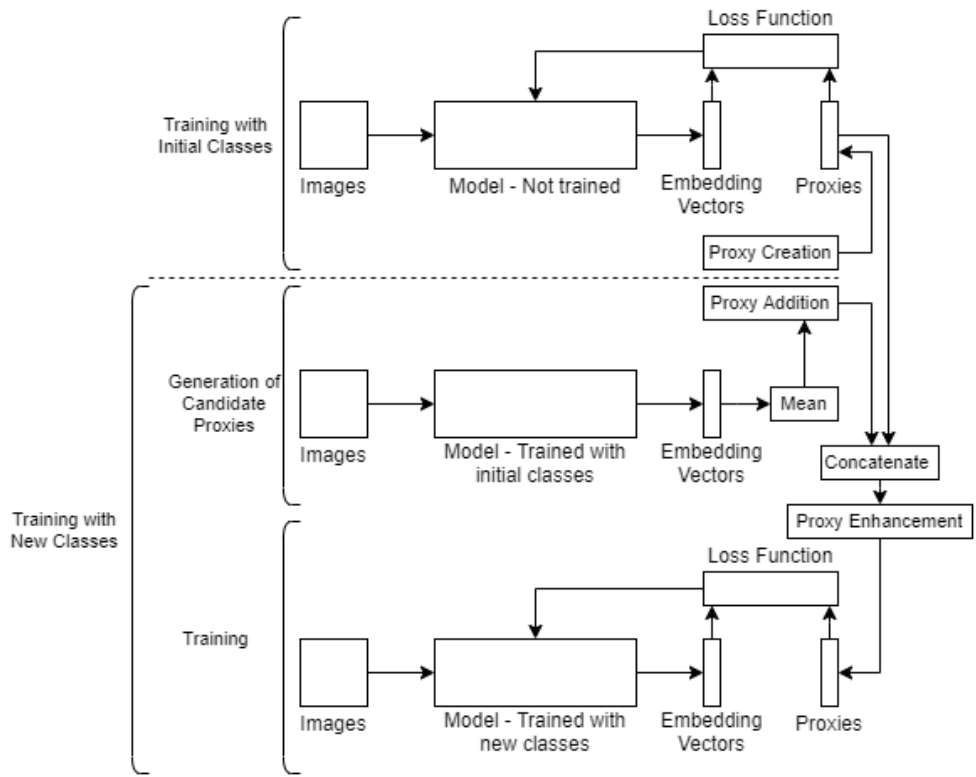

*Figure 10: Training processes with initial proxies and new proxies.*

### 5.2.1.1 Proxy Decomposition

After the proxies are defined randomly, a PyTorch linear layer class with one layer is defined where the randomly defined proxies are registered as parameters to be updated. This can be imagined as a fully connected layer in an artificial neural network where only the output layer is being updated. The loss function that we defined for decomposing the proxies uses the cosine distances calculated by the binary combination of each vector in the proxy array. After the cosine distances between each vector are calculated, they are tried to approach a zeros array (Figure 11).

$$Loss(s) = max(|s - Z|) + mean(|s - Z|) \qquad (1)$$

where:
s: Similarity matrix of proxies
Z: Zeros array

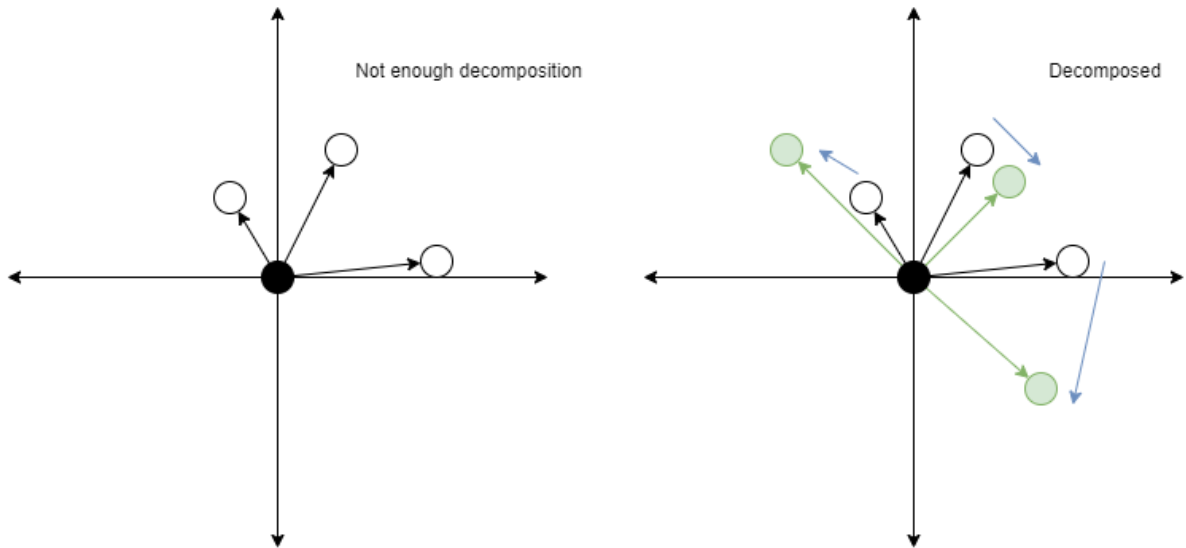

*Figure 11: Left graph shows the randomly created proxies. The right graph demonstrates the realignment of the initial proxies to be decomposed from each other. White circles indicate created proxies. Green circles indicate decomposed proxies.*

The goal of this operation is to decrease the mean and maximum value of cosine distances so that the proxies are going to get as sparse as possible. Any loss function that can backpropagate the error calculated by both maximum and mean similarity values can be used.

### 5.2.1.2 Proxy Addition

One of the technical requirements for the solution is to adjust the model to be able to identify new pill types while keeping the capability to predict the previous pill types. Because we use a proxy-based loss function, the behavior of our model differs from those trained with pair-based loss functions. Our model tries to map the input data to the pre-defined sparse representations of the classes that the input data belongs to. Therefore, our model cannot extract fine-grained features that can be used later to query during the inference. Hence, we developed a method to add new proxies to the existing proxy set.

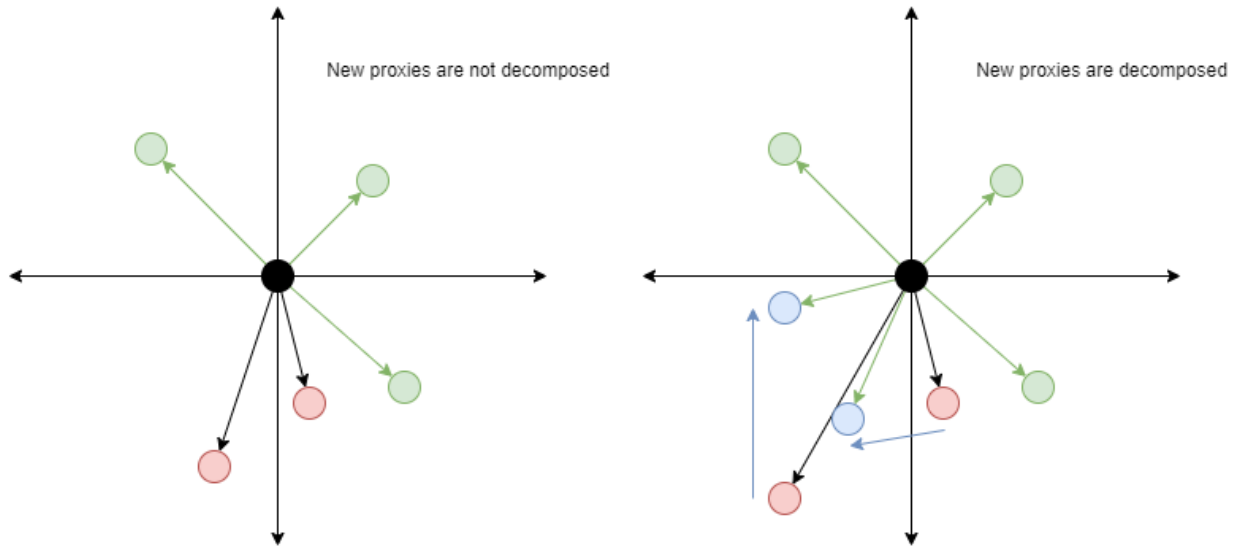

*Figure 12: Left graph shows the addition of new proxies (red circles). The right graph shows the realignment of the new proxies to be decomposed from each other and existing proxies. Green circles indicate previous proxies. Blue circles indicate the new positions of newly added proxies.*

Three important factors that should be considered concurrently to define the new proxies are determined:

1. New proxies should be decomposed from each other.
2. New proxies should be decomposed from the existing proxies.
3. New proxies should be as close as possible to the embedding vector that fits best to the embedding vectors that are generated from the images of new classes by the previously trained model.

According to the factors given above, a three-step process is designed to define the proxies (Figure 9):

1. By using the previously trained model, generate the embedding vectors of the images belonging to the new classes.
2. Calculate the candidate vectors of the embedding vectors of the images of the new classes that are generated in the first process by calculating the mean vector from the vectors of each class.
3. Optimize the candidate vectors by using gradient descend so that they are decomposed from each other while they are also decomposed from the existing proxies.

*5.2.1.3    Proxy Enhancement*

When new proxies are defined and decomposed, their sparsity is at its maximum. When new proxies are added and decomposed, they are decomposed from each other and existing proxies. Even though all proxies are decomposed in different sets, there is still room for more decomposition because they are not decomposed when all of them are registered as parameters to be updated at the same time. The enhancement is made to the new larger proxy set to maximize the sparsity of the proxies altogether.

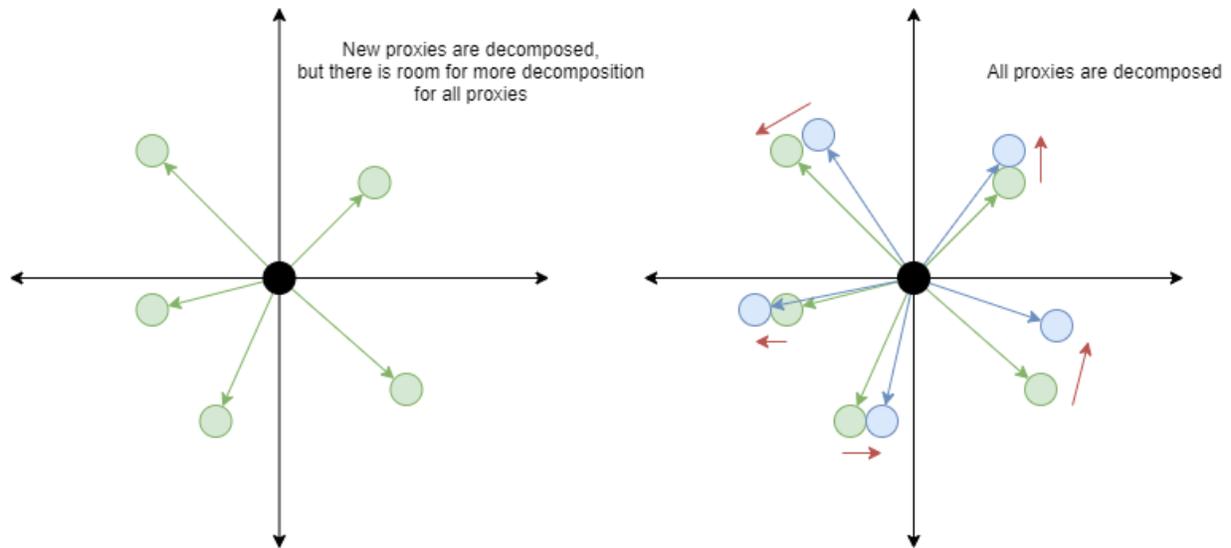

*Figure 13: Left graph shows the proxies after the addition process is done. Right graph shows the realignment of all proxies to be as decomposed as possible from each other.*

In this process, we reuse the same gradient descend method that is used in proxy decomposition. All proxies are defined as parameters to be updated so that the sparsity of the proxies will be at its maximum.

## 5.3 Classifier

During the development, the K-Nearest Neighbor (KNN) algorithm that was implemented in the repository of [3] was left untouched and used to classify the embedding vectors at the first stage of the research. Even though KNN is a broadly used algorithm that can achieve acceptable results, its dependency on the parameter of the number of nearest neighbors encouraged us to look for more innovative solutions for the classifier algorithm. In November 2020, the [4] is published. In the research, a new method to train the last layer of a convolutional neural network model is proposed. The method is used to train a fully connected layer in only one step by using the exact solution. There are several advantages of this method:

1. Because the weights are calculated just by using input and output rather than existing weights as gradient descend does, the challenges with weight initialization are avoided.
2. Because the weights are calculated by the Moore-Penrose inverse strategy, the most optimal pseudo-inverse of the input matrix is found. The use of the most optimal pseudo-inverse matrix to calculate the weights leads the weights to the most optimal values just in one step.
3. The fully connected layer that is trained with the exact solution does not depend on any parameter.

The use of the method using the exact solution is explained step by step below [4]:

1. A CNN model is trained with gradient descent for a couple of iterations.
2. The latent vectors are concatenated and then pseudoinverse of the concatenated vectors (input) is calculated by using the Moore-Penrose inverse strategy.
3. The input inverse matrix is multiplied with the expected output matrix. This operation generates the required weight matrix.
4. The obtained weight matrix is initialized to the last layer of the CNN model.

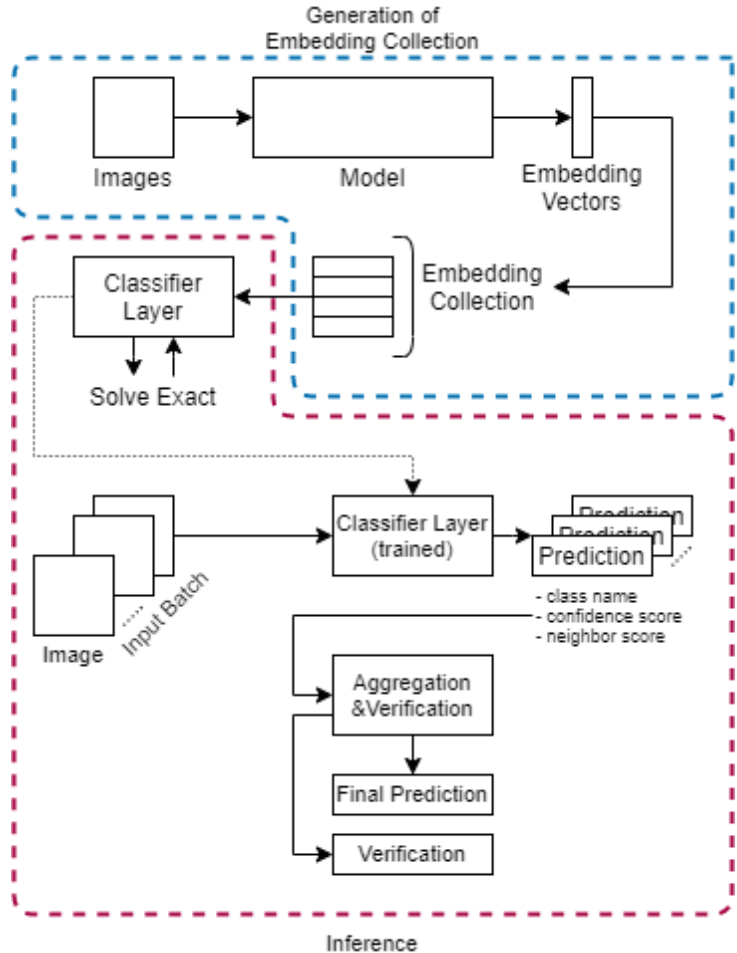

*Figure 14: Pill Identification Pipeline in Production-level*

This method has shown great achievements in entry-level benchmark datasets [4]. After investigating this research, we recognized that this method could be more useful if used in DML. The fully connected layer that was trained with the exact solution is called the "Solved Layer (SL)" later in this paper for convenience.

Because of its potential, we implemented this method and conducted experiments to compare the performance of the KNN and Solved Layer. During the experiments, the Solved Layer outperformed the KNN model (Table 7). Because of its performance and advantages, we employed Solved Layer to be used in our final solution as the classifier.

## 5.4 Aggregation and Verification Process

Besides a prediction for a single pill image, the input batch should be aggregated to narrow down the results to one single prediction. Therefore, we developed the aggregation/verification function to verify or deny the predictions. The confidence scores and number of occurrences of each predicted class are taken into account to design the aggregation/verification function. The steps of the aggregation/verification function are given below:

1. Get the prediction from the classifier, append them to a list.
2. Get the confidence score for each prediction, append them to a list.

3. Bind the prediction list and confidence score list together.
4. Sort the list of predictions with respect to their confidence scores.
5. Get the pill type (and its highest confidence score) that occurs most in the first 10 elements of the sorted list of predictions created in step 4.
6. Get the pill type and confidence score of the pill type that has the highest confidence score in the first 10 elements of the sorted list of predictions created in step 4.
7. Assign "True" to the parameter "condition 1" if the confidence score of the pill type found in step 5 is higher than the minimum threshold of 0.87. Assign "False" to the parameter "condition 1", if-else. The minimum threshold is determined after we observed the confidence scores manually in the experiments.
8. Assign "True" to the parameter "condition 2" if the pill type that has the highest confidence score and the pill type that has the highest occurrence are the same. Otherwise, apply the condition in the next step.
9. Assign "True" to the parameter "condition 2" if the difference between the confidence scores of the pill type that has the highest confidence and the pill type that occurs most is less than or equal to 0.1. Assign "False" to the parameter "condition 2". If the difference is more than 0.1, assign "False" to the parameter "condition 2".
10. Apply the logic operation "Condition 1 AND Condition 2" and assign the result to the parameter "condition".
11. Determine the most occurrent pill type as the prediction. If the parameter "condition" is "True", the prediction is verified. If the parameter "condition" is "False", the prediction is not verified.

By the process above, the solution that we developed can achieve 0 false positives while maximizing the metric "Ratio of Verified Predictions" (See Section 6.1).

# 6. Experiments

Experiments were conducted with two datasets that we prepared for this project (See Section 3). For the implementation of the code for the experiments and overall solution, we forked the Proxy Anchor Loss repository [3] and built our solution on top of it.

Instead of making experiments for all combinations of backbone architectures and loss functions, we separated the experiments for different purposes. Therefore, several different experiments were applied to make decisions for methods, models, and loss functions.

## 6.1 Metrics and Methods for Evaluation

To evaluate the performance of the model, "Model Performance Test", "Single Vial-based Test", "Multiple Vial-based Test" tests were conducted.

### 6.1.1 Model Performance Test

This test was made with an evaluation set that was separate from the training and test set (Figure 8). The test was done during the training of the model after each epoch. It let us monitor the performance of the model from a basic perspective and, it showed the approximate accuracy that the model can achieve in final results. The metric that we observed in this test is named "Micro accuracy". The micro accuracy metric is the same as Recall@K in [3]. An increment in the micro accuracy represents the increment in the model's performance at the production level. The classifier that we used in this test is KNN and, models were trained for 20 epochs.

*6.1.2   Single Vial Test*

This test was made with a test set that was separate from both the training set and evaluation set. This test represents the accuracy of the model at the pre-production level. In the test set, one input sample refers to a batch of single pill images that are segmented from one pill vial. Therefore, the model not only gives a prediction for each pill but also aggregates the predictions of the pills in the input sample. During the aggregation process, the function that executes the aggregation also verifies the final prediction. If the final prediction is verified, it means that the model is sure about the correctness of the prediction. If the final prediction is not verified, it means that the prediction is not reliable, therefore more information (another batch of images that belongs to the same pill type) should be collected and fed into the model. In the test, we used two different metrics which are "Accuracy of All Predictions" and "Ratio of Verified Predictions". "Accuracy of All Predictions" shows the ratio of accurately predicted samples to total samples. "Ratio of Verified Predictions" is the ratio of the number of verified predictions for all samples to the number of total predictions for all samples. "Ratio of verified predictions" has to be less than or equal to the "Accuracy of All Predictions" because only the samples that are predicted correctly can be verified by the model. Besides, we observed the metric "Accuracy of Verified Predictions" to see if the aggregation/verification function works properly.

*6.1.3   Multiple Vial Test*

This test is almost identical to the single vial test. The only difference between the single vial test and the multiple vial test is the input sample. At the production level, the input sample that comes from the VeriMedi device is not only a single whole vial image. The input sample at the production level contains images of the same vial with seven different light conditions. The test was made to observe the performance of the model at the production level. Hence, the aggregation/verification function was applied to the predictions of single pill images that were segmented from seven different poses rather than the predictions of single pill images that are segmented from an image having one light condition.

## 6.2 Experiment to Choose the Background Type in Pill Segmentation

Blurred (normalized box filter having the size of 10), gray (pixel value 128 for RGB channels), and bounding box (no background change) backgrounds are applied to segmented single pill images from the ShakeNet dataset. A pill identification model (see 2.4) is trained and the accuracy was calculated as a ratio of correct predictions against the total number of pills (Model Performance Test, See Section 6.1.1). The metric that we observe in this test is micro accuracy (section 6.1.1).

According to the test results (Figure 15), the blurred background outperforms two other background types slightly. Therefore, we have chosen the blurred background as the background type to be applied after the pills are segmented.

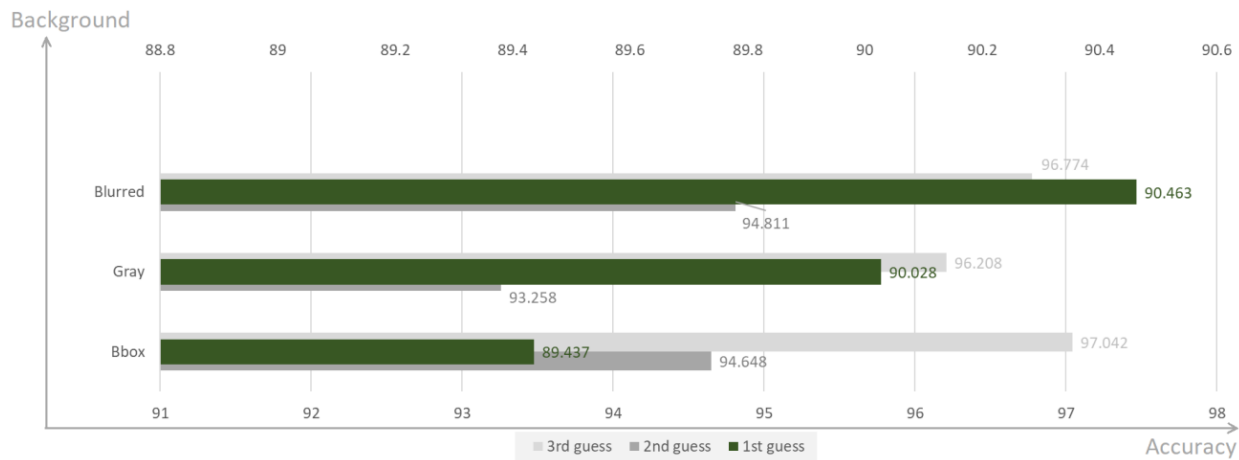

*Figure 15: Assessment of different masking types. "Model Performance Test" is applied to generate the results. Blurred pills achieved the highest score.*

## 6.3 Experiment to Choose Backbone Architecture

The backbone architectures that we compare in this experiment are GoogleNet [18], BN-Inception [19], ResNet-18, ResNet-34, ResNet-50, ResNet-101 [11]. The loss function that was used to compare the performance of backbone architectures is PAL [3]. The test that we applied is the single vial test and the metrics that we observed in this experiment are the ratio of verified predictions and accuracy of all predictions. Besides, we also demonstrated the accuracy of verified predictions in the metrics so that we could observe if the aggregation/verification process works as expected. The results are shown in Table 2.

| Dataset | Model Architecture | Proxy Decomposition | Accuracy of Verified Predictions | Ratio of Verified Predictions | Accuracy of All Predictions |
|---|---|---|---|---|---|
| ShakeNet | BN-Inception | 0 | 100.00 | 84.29 | 91.13 |
|  | GoogleNet | 0 | 100.00 | 87.86 | 93.84 |
|  | ResNet-18 | 0 | 100.00 | 91.67 | 97.04 |
|  | ResNet-34 | 0 | 100.00 | **96.19 (1)** | **99.51** |
|  | ResNet-34 | 1 | 100.00 | **94.29 (2)** | **98.29** |
|  | ResNet-50 | 0 | 100.00 | **94.29 (3)** | **97.78** |
|  | ResNet-50 | 1 | 100.00 | 88.33 | 93.60 |
|  | ResNet-101 | 0 | 100.00 | 90.24 | 93.84 |
|  | ResNet-101 | 1 | 100.00 | 79.76 | 86.21 |
| SyntheticNet | BN-Inception | 0 | 100.00 | 77.00 | 78.00 |
|  | GoogleNet | 0 | 100.00 | 71.00 | 73.00 |
|  | ResNet-18 | 0 | 100.00 | 75.50 | 77.50 |
|  | ResNet-34 | 0 | 100.00 | **81.00 (3)** | **81.00** |
|  | ResNet-34 | 1 | 100.00 | **85.50 (1)** | **85.50** |
|  | ResNet-50 | 0 | 100.00 | 78.50 | 79.00 |
|  | ResNet-50 | 1 | 100.00 | 80.00 | 80.00 |
|  | ResNet-101 | 0 | 100.00 | **82.00 (2)** | **83.50** |
|  | ResNet-101 | 1 | 100.00 | 80.00 | 80.50 |

*Table 2: The results of the experiment that is conducted to evaluate and choose the backbone architecture for the DML model. Numbers in brackets indicate the rank of success. 0 and 1 in the Proxy Decomposition column show the status of the Proxy Decomposition as False and True respectively.*

In the ShakeNet dataset, ResNet-34 architecture that is trained with undecomposed proxies outperformed all other architectures in all metrics. The ResNet-34 architecture trained with decomposed proxies achieved the second-best score for accuracy of all predictions, while it shares the second rank with ResNet-50 trained with undecomposed proxies in the ratio of verified predictions.

In the SyntheticNet dataset, ResNet-34 with decomposed proxies outperformed all other architectures in all metrics. ResNet-101 architecture trained with undecomposed proxies achieved the second-best scores for all metrics. ResNet-34 with decomposed proxies achieved the third-best scores for all metrics.

According to the experiment results, we decided to continue with ResNet-34 architecture.

## 6.4 Experiment to Choose Loss Function

For this experiment, we kept the backbone architecture the same for all loss functions to be able to compare them. The backbone architecture that we used for the experiment is ResNet-34. The loss functions that we compared are Contrastive Loss [20], Triplet Loss [21], Multi-Similarity Loss [22], Proxy-NCA Loss [23], and PAL [3]. The main metric that we observed for this experiment is the ratio of verified predictions. To compare the performance of the models with different loss functions, we used the PAL repository [3] that uses the PyTorch Metric Learning library [24] to implement the other loss functions.

| Dataset | Loss Function | Proxy Decomposition | Accuracy of Verified Predictions | Ratio of Verified Predictions | Accuracy of All Predictions |
|---|---|---|---|---|---|
| ShakeNet | Contrastive | 0 | 100.00 | 86.67 | 96.80 |
| | Triplet | 0 | 100.00 | 96.30 | 90.95 |
| | MS | 0 | 100.00 | 90.00 | 96.06 |
| | Proxy NCA | 0 | 100.00 | 86.90 | 95.57 |
| | Proxy Anchor | 0 | 100.00 | **96.19 (1)** | **99.51** |
| | Proxy Anchor | 1 | 100.00 | **94.05 (2)** | **97.78** |
| SyntheticNet | Contrastive | 0 | 100.00 | 68.50 | 69.00 |
| | Triplet | 0 | 100.00 | 74.50 | 77.50 |
| | MS | 0 | 100.00 | 74.00 | 77.00 |
| | Proxy NCA | 0 | 100.00 | 77.00 | 77.00 |
| | Proxy Anchor | 0 | 100.00 | **81.00 (2)** | **81.00** |
| | Proxy Anchor | 1 | 100.00 | **85.50 (1)** | **85.50** |

*Table 3: The results of the experiment that is conducted to evaluate and choose the loss function for the DML model. Numbers in brackets indicate the rank of success. PAL with undecomposed proxies has achieved the highest accuracy in ShakeNet, while PAL with decomposed proxies has achieved the highest accuracy in SyntheticNet. Numbers in brackets indicate the rank of success.*

According to the results (Table 3), it can be seen that PAL achieved the highest scores for both metrics and both datasets. While the decomposition operation increased the ratio of verified predictions for the SyntheticNet dataset by 4.5%, it decreased the same metric for ShakeNet by 2.14%. Because the number of pill types in ShakeNet was low, it was not expected that the proxy decomposition would make a positive impact in ShakeNet. In contrast, the proxy decomposition helped the model to increase the ratio of verified predictions and accuracy in the SyntheticNet. The main reason for it is that SyntheticNet had 5 times the pill types that the ShakeNet had. Therefore, the decomposition process becomes more required for the SyntheticNet.

Even though the decomposition helped the model to perform better only in the SyntheticNet dataset, it is still required to use if we want our model to be able to learn continually. Therefore we decided to use PAL using Proxy Decomposition.

## 6.5 Demonstration of Proxy Operations

To demonstrate and create an intuition to understand how the optimization is working, a Python script was prepared. The script creates 4 proxies with embedding vectors having size 2 and optimizes them to be decomposed from each other (Proxy Creation). Then, it adds 2 more proxies and optimizes them to be decomposed from each other and initial proxies (Proxy Addition). Finally, the script optimizes the final six proxies to be decomposed from each other (Proxy Enhancement). In conclusion, the similarity scores of the proxies should be as small as possible. The maximum similarity score between the proxies can be found in Table 4.

| Proxy Operation | Maximum Similarity | |
|---|---|---|
| | Before proxy operation | After proxy operation |
| **Proxy Creation** | 0.8961 | 0.7074 |
| **Proxy Addition** | 0.9993 | 0.9252 |
| **Proxy Enhancement** | 0.9259 | 0.8689 |

*Table 4: Demonstration of maximum similarity scores between defined proxies. The before refers to the condition that the relevant proxy operation is not applied while after refers to the condition that the relevant proxy operation is applied.*

Table 4 shows, that the proxies, which were initialized with normal distribution have a maximum similarity score of 0.8961, which means that the sparsity of the proxies is low, and this may cause the model to make mispredictions even after converging to a low error rate. During the proxy creation, we decompose the initial proxies to increase the sparsity of the proxies. The maximum similarity score of the proxies after the decomposition is 0.7074, which is significantly lower than the maximum similarity score of the proxies before decomposition.

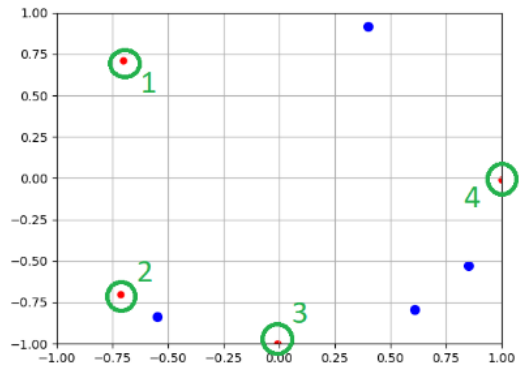

*Figure 16: Demonstration of undecomposed and decomposed proxies in the proxy creation process. Blue points are undecomposed proxies, red points that are encircled with green are decomposed proxies. All proxies are normalized to fit into the range between -1 and 1.*

After the proxy creation, new proxies were added, which corresponds to the addition of new classes that the model can predict. When new proxies were added with the random initialization, the maximum similarity between the proxies reached 0.9993, which means that one or more of the new proxies had a pretty similar orientation to one of the existing proxies. When the decomposition was applied during the proxy addition process, only the new proxies were registered as parameters to be updated, while old proxies were kept the same (Figure 17).

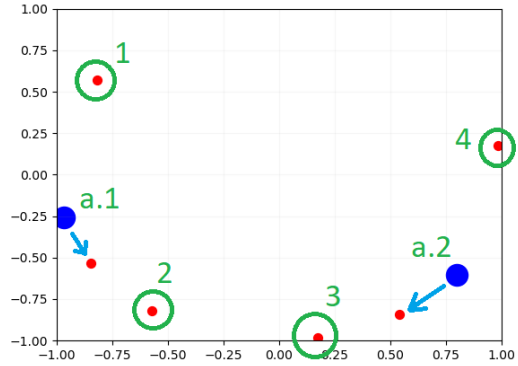

*Figure 17: Demonstration of undecomposed and decomposed proxies in the proxy addition process. Green circles 1, 2, 3, 4 indicate previous proxies, a.1 and a.2 are the initial positions of new proxies, arrows show the movement of new proxies. All proxies are normalized to fit into the range between -1 and 1.*

As was mentioned in the previous paragraph, only the new proxies were optimized to lower the maximum similarity score in the proxy addition process. The reason to have the previous proxies unchanged is to keep the training complexity of the model low and avoid retraining the model from scratch, which would be necessary if the existing proxies were moved as well. After the new proxies were optimized, enhancement of the proxies was applied. As a result, the maximum similarity score of the proxies was decreased from 0.9259 to 0.8689. Thus, with the proxy optimization process, the sparsity of the proxies was increased without a significant effect on the training complexity of the model (Figure 18).

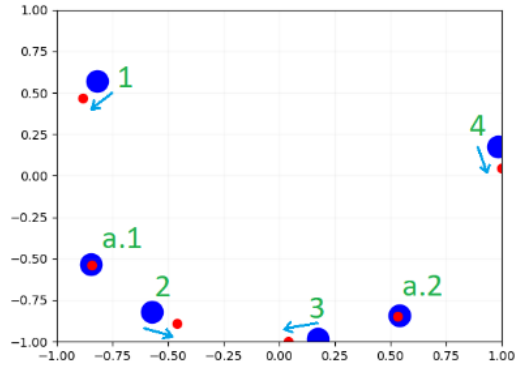

*Figure 18: Demonstration of Proxy Enhancement. Blue points indicate previous positions of proxies, red points indicate new positions of the proxies. All proxies are normalized to fit into the range between -1 and 1.*

## 6.6 Experiments to Choose Classification Algorithm

We conducted these tests to compare the performance of KNN and Solved Layer. ResNet-50 [11] backbone and PAL were used to train the model. The test type is "Pre-production Level Single Vial-based Test". As can be seen in Table 5, Solved Layer has not made a considerable improvement in the test with the ShakeNet dataset. On the other hand, Solved Layer increased the recall with verification by 5.5 percent compared to the KNN in the test with SyntheticNet dataset. Therefore, it was decided to use the "Solved Layer" as the classifier of the identification solution.

| Metrics | Accuracy of All Predictions | | Recall w/o Verification | | Accuracy of Verified Predictions | | Ratio of Verified Predictions | |
|---|---|---|---|---|---|---|---|---|
| Algorithm | KNN | SL | KNN | SL | KNN | SL | KNN | SL |
| ShakeNet | 96.55% | 97.29% | 100.00% | 100.00% | 100.00% | 100.00% | 91.43% | 91.19% |
| SyntheticNet | 79.00% | 78.50% | 100.00% | 100.00% | 100.00% | 100.00% | 72.50% | 78.00% |

Table 5: The test is executed to compare the performance of the models having different classification algorithms. The test type is 2 (section 8.1). ResNet-50 backbone is used for the tests.

# 7. Results

## 7.1 The Performance of the Final Identification Solution

The final pill identification solution includes the full cycle of image pre-processing, segmentation, and identification for the images acquired by the VeriMedi device or ShakeNet. Although it was not feasible to simulate different light conditions and variety of pills distribution for the images of SyntheticNet dataset as it appears for ShakeNet, the results of the tests made with SyntheticNet are still included in the assessment (Table 6). The test that we applied here was the multiple vial test (See Section 6.1.3).

According to the test results (Table 6), the solution achieved 100% accuracy for all predictions while it verified 96.66% of predictions with the ShakeNet dataset. The results are higher than the results in the tests that were made with single-vial images (Table 2, Table 3). The performance of the solution has achieved 89% accuracy for all predictions with the SyntheticNet, which also indicates improved performance in comparison with the tests that were made with single-vial images (Table 2, Table 3). This proves that higher accuracy is achieved when single-pill images coming from different poses are used together to make a prediction.

| Dataset | Accuracy of Verified Predictions | Ratio of Verified Predictions | Accuracy of All Predictions |
|---|---|---|---|
| ShakeNet | 100.00 | **96.66** | **100.00** |
| SyntheticNet | 100.00 | 89.00 | 89.00 |

Table 6: The test results of the final identification solution using single-pill images coming from multiple poses.

## 7.2 The Performance of Identification Solution on Avoidance of Unknown Classes

To assess the capability of the identification solution to avoid giving verified predictions for unknown classes, we conducted tests where the results are displayed in Table 7. In the test, we trained the model with one half (50%) of the dataset and evaluated the performance of the model with the other half (50%) of the dataset. The accuracy of verified predictions and accuracy of all predictions are always 0 because the classes of the first half (collection) and the second half (queried) are completely different. On the other hand, the model is not supposed to verify the given predictions because none of them were true. Therefore, the metric that we observed in this test is the ratio of verified predictions, and the expected value for the metric is 0. "Dataset – Train" is the split that was used to train the model. "Dataset – Collection" is the split that was used to generate reference embedding vectors. "Dataset – Test" is the split that was used to test the model (Table 7). The overall process that we followed to test the performance of the model to avoid unknown classes is given below:

- Divide the dataset into two different parts where each part contains one half of the whole dataset.
- Train the model with the first half of the dataset for 20 epochs.
- Generate embedding vector collection with the first half of the dataset.
- Test the solution with the second half of the dataset and save the results.
- Train the model with the second half of the dataset for 20 epochs.
- Generate embedding vector collection with the second half of the dataset.
- Test the solution with the first half of the dataset and save the results.

According to the results in Table 7, the ratio of verified predictions is 84.29% and 74.28% for both the first and second halves of the ShakeNet dataset. This shows that model was not capable to avoid unknown classes. Even though the embedding vector collection does not have the unknown classes, the results show that the model still verifies the wrong predictions.

For both parts of the SyntheticNet dataset, the ratio of the verified predictions is low compared to ShakeNet results. But the values are still high because the expected outcome is 0. This shows that the model was capable of avoiding 20.72% (100.00-79.28) of the unknown classes in the ShakeNet dataset and was capable of avoiding 79% (100.00-21.00) of the unknown classes in the SyntheticNet dataset. These scores show that the model is not successful in avoiding unknown classes. It is because the model is associating the features that it learned to predict a class that it hasn't learned before. The predictions that it makes have a confidence score that is high enough to verify the given predictions.

| Dataset - Train | Dataset - Collection | Dataset - Test | Ratio of Verified Predictions (%) | Accuracy of All Predictions (%) | Average Ratio of Verified Predictions (%) |
|---|---|---|---|---|---|
| ShakeNet (1-10) | ShakeNet (1-10) | ShakeNet (11-20) | 84.29 | 0.00 | 79.28 |
| ShakeNet (11-20) | ShakeNet (11-20) | ShakeNet (1-10) | 74.28 | 0.00 | |
| SyntheticNet (1-50) | SyntheticNet (51-100) | SyntheticNet (51-100) | 26.00 | 0.00 | 21.00 |
| SyntheticNet (51-100) | SyntheticNet (1-50) | SyntheticNet (1-50) | 16.00 | 0.00 | |

*Table 7: The test results of the final solution on unknown classes. The Performance of the Identification Solution of new classes with Proxy Addition*

## 7.3 The Performance of the Identification Solution with New Classes

This test helped us observe the quality of the embedding vector collection that was generated by the model for unknown classes which were not introduced to the model during training. The experiment also helped us compare the performance of the model for both known and unknown classes before and after training with the unknown classes. To test the performance of the model with the addition of new classes (proxy addition), we followed the steps given below:

- Divide the dataset into two different parts where each part contains one half of the whole dataset.
- Train the model with the first half of the dataset for 20 epochs.
- Generate embedding vector collection with the first half of the dataset.
- Test the solution with the first half of the dataset and save the results.
- Generate the embedding vector collection with the second half of the dataset. (Reminder: The model is not trained with the second half yet. The model is trained with the only first half of the dataset.)

- Test the solution with the second half of the dataset and save the results.
- Train the model (pre-trained with the first half of the dataset) with the second half of the dataset for 20 epochs.
- Generate embedding vector collection with the first half of the dataset.
- Test the solution with the first half of the dataset and save the results.
- Generate the embedding vector collection with the second half of the dataset.
- Test the solution with the second half of the dataset and save the results.

| Dataset - Train | Dataset - Collection | Dataset - Test | Accuracy of Verified Predictions (%) | Ratio of Verified Predictions (%) | Accuracy of All Predictions (%) |
|---|---|---|---|---|---|
| ShakeNet (1-10) | ShakeNet (1-10) | ShakeNet (1-10) | 100.00 | 91.43 | 97.96 |
| ShakeNet (1-10) | ShakeNet (11-20) | ShakeNet (11-20) | 100.00 | 93.33 | 96.67 |
| ShakeNet (11-20) | ShakeNet (1-10) | ShakeNet (1-10) | 100.00 | 86.67 | 94.38 |
| ShakeNet (11-20) | ShakeNet (11-20) | ShakeNet (11-20) | 100.00 | 98.57 | 98.57 |
| SyntheticNet (1-50) | SyntheticNet (1-50) | SyntheticNet (1-50) | 100.00 | 83.00 | 84.00 |
| SyntheticNet (1-50) | SyntheticNet (51-100) | SyntheticNet (51-100) | 100.00 | 84.00 | 85.00 |
| SyntheticNet (51-100) | SyntheticNet (1-50) | SyntheticNet (1-50) | 100.00 | 86.00 | 87.00 |
| SyntheticNet (51-100) | SyntheticNet (51-100) | SyntheticNet (51-100) | 100.00 | 93.00 | 93.00 |

*Table 8: Performance of the final solution using the Proxy Addition method*

The ratio of verified predictions and accuracy of all predictions decreased by 4.76% and 3.58% respectively for the first half of the ShakeNet dataset when the model was retrained with only the second half of the dataset. In contrast, the ratio of verified predictions and accuracy of all predictions increased 5.27% and 1.9% respectively for the second half of the ShakeNet dataset when the model was retrained with the second half of the dataset.

The ratio of verified predictions and accuracy of all predictions increased 3% for the first half of the SyntheticNet dataset when the model was retrained with only the second half of the SyntheticNet dataset. The ratio of verified predictions and accuracy of all predictions increased 9% and 8% respectively for the second half of the SyntheticNet dataset when the model was retrained only with the second half of the dataset.

The results show that proxy addition and retraining only with the new proxies (classes) increases the performance of the model on the second halves of the datasets. It means that when the model is retrained with the new classes, it is able to learn new features and associate all features for identifying the new classes. In ShakeNet, the first half of the dataset showed lower performance when the model was retrained with the second half. In SyntheticNet, retraining only with the second half increased not only the performance of the second half of the dataset but also the performance of the first half of the dataset. This shows that the proxy addition process is successful enough to be used when the new pill types are required to be predicted. Even though we haven't investigated why the model's performance gets better for the first half of the SyntheticNet dataset when it is retrained only with the second half, it is likely that the model learns new features and uses them to better distinguish between the old classes as well.

## 8. Discussions and Further Research

As shown in Section 7.2, the solution is not capable of avoiding unknown classes. Therefore, it is necessary to research this problem.

To avoid unknown classes, we designed and partially implemented Feature-Based Proxy Anchor Loss. Feature-based Proxy Anchor Loss (FBPAL) is similar to PAL with Proxy Operations. FBPAL works with fragments of the embedding vector that is generated by the DML model.

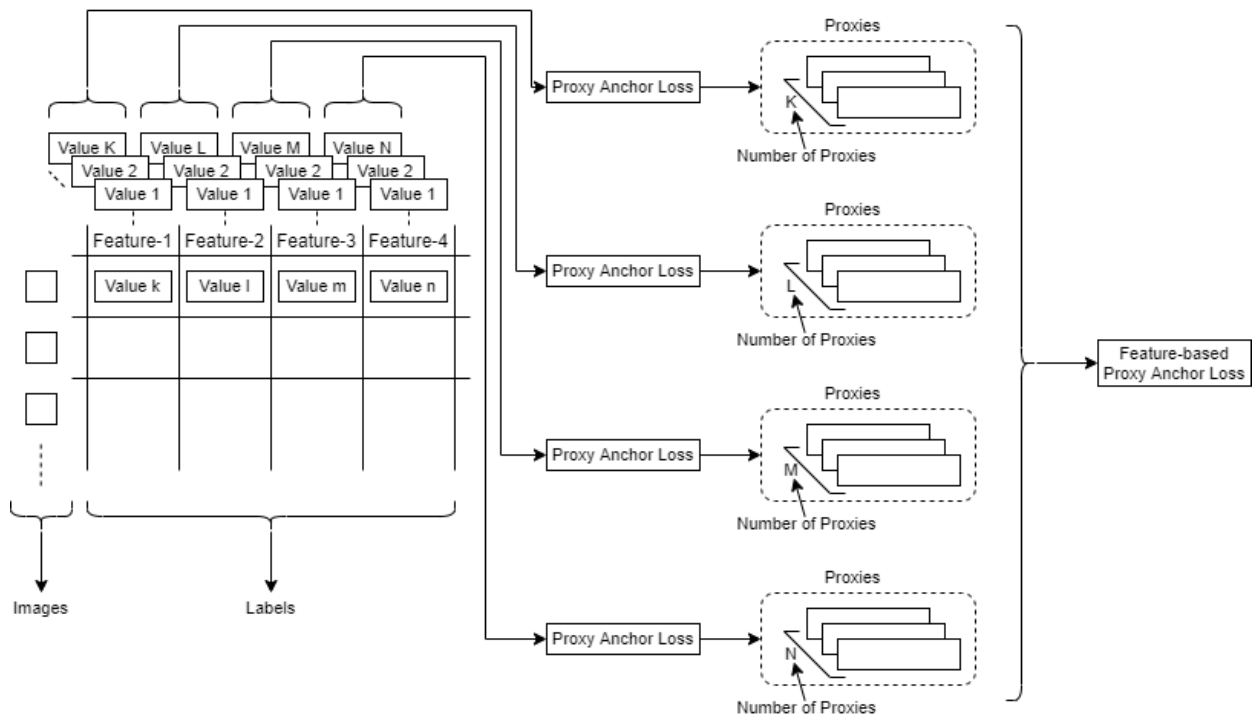

*Figure 19: Labeling process and definition of Feature-based Proxy Anchor Loss. Feature-based Proxy Anchor Loss is composed of several Proxy Anchor Loss functions where each of them has a different number of proxies for different features. The number of values a feature can take defines the number of proxies for the defined Proxy Anchor Loss instance for that feature.*

For example, a DML model generates an embedding vector with a size of 512. FBPAL imports the embedding vector and split it into embedding vector pieces with a size of 128 (4 pieces). On the other hand, the images in the dataset are labeled with their features rather than the pill types (labeling with features: brown, round, black imprint, characters "a, d, v, i, l", labeling with pill type: Advil 200mg). After the images are labeled with features, similarities between the possible values for each feature are calculated and a similarity matrix is created. Then, proxies of the features are created and decomposition is applied on the created proxies to fit the similarity matrix of the features so that the similarity values of the proxies between each other will be close to the values in the similarity matrix of feature labels. After that, FBPAL is used in the same way that the Proxy Anchor Loss is used to train the model. The model that is trained with FBPAL is going to learn the high-level features rather than pill types (Figure 19). Because this method would lead the model to learn the high-level features in the image, it might be able to distinguish unknown classes from known classes with respect to the features that the model has generated. Besides, this method can also help train models that can satisfy the principles of the Explainable AI [25] because the models are trained to learn high-level human-readable features before they make a classification. Further research is required on this subject.

Proxy-based DML models are pretty successful for representing information as can be seen in the results from [3] and this research. While the next stage FBPAL can help us create Explainable AI [25] systems, there is another approach that can support Explainable AI [25] principles. The method is another version of PAL and FBPAL. The proxies in PAL are multi-dimensional vectors where each class/concept is tried to be fit to a specific predefined proxy. In FBPAL, it is proposed that the predefined proxies can be used for representing features rather than classes and concepts. In the new method, the features are not predefined. The proxies are defined previously as is in the PAL but the embedding vectors that are generated by the model are not directly fit into those proxies. Instead, an embedding vector that is generated by the model is split into embedding vector pieces as in FBPAL and these embedding vector pieces are summed. Another vector is acquired by the summation of embedding vector pieces. This vector is tried to fit the predefined decomposed proxies. In this case, the size of proxies should be the same size as embedding vector pieces rather than the size of the whole embedding vector (Figure 21). The embedding vector pieces are treated as feature vectors and the concept/class is accepted as the f]inal product of the summation of feature vectors (Figure 20). This method is named "Proxy Anchor Loss with Feature Summation (PALFS)".

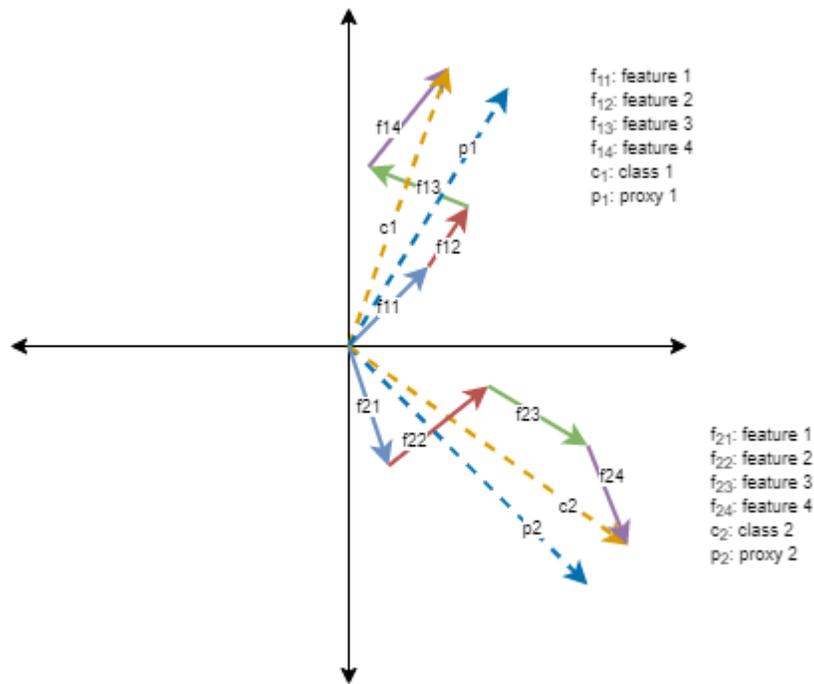

*Figure 20: An Example for PALFS. Feature vectors (f11, f12, f13, f14) are summed and the concept/class vector (c1) is acquired, the proxy of c1 is p1. Feature vectors (f21, f22, f23, f24) are summed and the concept/class vector (c2) is acquired, the proxy of c2 is p2.*

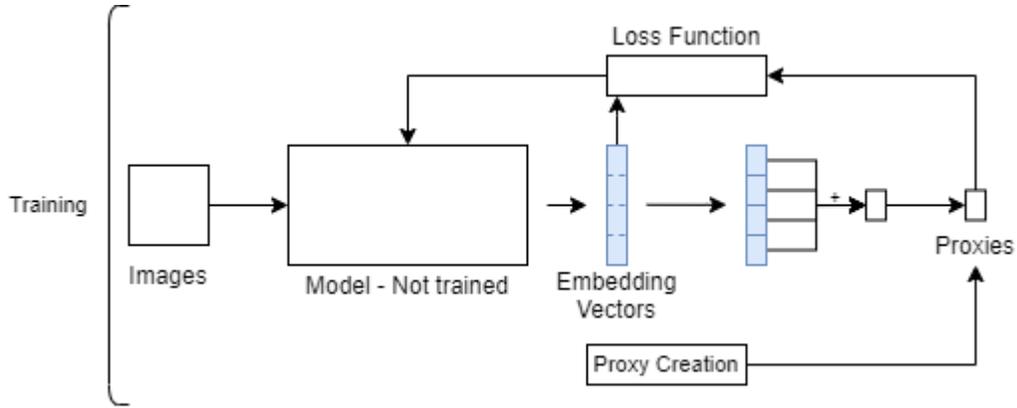

*Figure 21: Training process with PALFS. Embedding vectors are generated by the DML model, generated embedding vectors are split into embedding vector pieces that have the size of predefined feature size. These vector pieces are summed and fed to PAL to calculate the loss.*

To evaluate the capability of PALFS, we conducted a small experiment. In the experiment, we used PALFS with ResNet-34 architecture for both ShakeNet and SyntheticNet datasets. According to the test results, PALFS outperformed the proposed solution (Table 3) (DML model with ResNet-34 architecture that was trained with enhanced PAL function.) in "Accuracy of All Predictions", while the proposed solution outperformed PALFS in "Ratio of Verified Predictions" (Table 9). According to the results, PALFS is giving promising results, but it is still not capable of outperforming the proposed solution in the verified predictions. Therefore, further research (especially in the aggregation/verification function of the PALFS solution) is required.

| Dataset | Loss | Accuracy of Verified Predictions | Ratio of Verified Predictions | Accuracy of All Predictions |
|---|---|---|---|---|
| ShakeNet | ProxyAnchor | 100.00 | **94.05** | 97.78 |
| ShakeNet | PALFS | 100.00 | 91.43 | **97.54** |
| SyntheticNet | ProxyAnchor | 100.00 | **85.50** | 85.50 |
| SyntheticNet | PALFS | 100.00 | 82.00 | **88.50** |

*Table 9: The performance comparison for proposed solution and PALFS.*

FBPAL and PALFS work with features rather than classes directly. In FBPAL, features are strongly controlled. In PALFS, features are summed before they are normalized so that they form the vector of the concept/class. The summation is made without any weight on any feature. However, a concept can have a different weight for each feature that it contains. Therefore, a weighted sum rather than a direct sum could create a difference. We conducted the PALFS experiment also with weighted sum of the features (Table 10). According to the experiment results, the accuracy of all predictions has increased slightly for the ShakeNet dataset. In the results of the SyntheticNet dataset, there is a drastic decrease in both metrics.

| Dataset | Loss | Accuracy of Verified Predictions | Ratio of Verified Predictions | Accuracy of All Predictions |
|---|---|---|---|---|
| ShakeNet | ProxyAnchor | 100.00 | 94.05 | 97.78 |
| ShakeNet | PALFS -weighted sum | 99.49 | **94.29** | **98.28** |
| SyntheticNet | ProxyAnchor | 100.00 | **85.50** | **85.50** |
| SyntheticNet | PALFS -weighted sum | 99.28 | 69.50 | 75.50 |

*Table 10: The performance comparison for the proposed solution and PALFS with weighted sum.*

This shows that the PALFS with weighted sum also has potential but requires further research.

Another future work is to apply the identification solution to other problem domains such as product identification and face identification.

Lastly, the use of proxies to define a class rather than a binary vector may change the way we approach deep learning in general. The segmentation models using pixel to pixel approach may leverage proxies to represent the classes. If the proxy operations that are proposed in this paper are also used with the proxy-based segmentation models, the number of object types that can be segmented can be enlarged easily.

We believe that the plasticity in the human brain is what we need to make breakthroughs in deep learning. And, we believe that the Proxy Operation methods that we have told and proposed in this research are the first step to it.

## 9. Conclusion

- A complete solution that can do segmentation and identification for pill images was designed and developed.
- The developed segmentation solution can segment the pills and also capture the imprint and engraving information from the pills and the vial so that noisy pills and informative pills could be detected.
- The DML model that is used to generate embedding vectors is capable of learning continually thanks to the Proxy Operations that are proposed in this research.
- The final solution is capable of reaching 100% accuracy and 96.66% verification for the ShakeNet dataset that contains production-level images from the VeriMedi device.
- The pill identification model is capable of learning from the limited number of images per class.
- While the segmentation solution is specific to pills, the identification solution can be reused for classification tasks from other problem domains.
- Two unique datasets are created to work with vial-based images.
- The proposed methods in Proxy Operations create a new approach in deep learning where we accept information as a set of multi-dimensional vectors rather than a binary representation.

Even though the solution has several capabilities, it does not have one crucial capability which is to avoid unknown classes. Therefore, further research on this area is required.

## 10. Acknowledgments

Firstly, we would like to thank Ethan R. Bischoff from Accenture USA who supported and funded this project. Ethan's leadership and wide vision as well as financial support carried us during the project. Secondly, thanks to Mark H. Olson, Alexander Hoppe, Akash Idnani, and all other MindTribe team members for their great work and collaboration on the project. Thirdly, we would like to thank Olugbenga A. Omoteso for his management and support for the project. We also would like to thank Solvita Berzisa for her comments and suggestions in the preparation of this paper. Lastly, we would like to thank Thomas Jude Schmit for proofreading.